\documentclass{article}

\usepackage[preprint,nonatbib]{neurips_2024}

\usepackage[utf8]{inputenc} 
\usepackage[T1]{fontenc}    
\usepackage{hyperref}       
\usepackage{url}            
\usepackage{booktabs}       
\usepackage{amsfonts}       
\usepackage{nicefrac}       
\usepackage{microtype}      

\usepackage[dvipsnames,table]{xcolor}
\usepackage[numbers]{natbib}
\usepackage{multirow}
\usepackage{graphicx}
\usepackage{wrapfig}
\usepackage{algorithm}
\usepackage{algpseudocode}
\usepackage{makecell}

\usepackage{amsmath,amsfonts,bm}

\def\eqref#1{equation~\ref{#1}}

\def\1{\bm{1}}

\DeclareMathAlphabet{\mathsfit}{\encodingdefault}{\sfdefault}{m}{sl}
\SetMathAlphabet{\mathsfit}{bold}{\encodingdefault}{\sfdefault}{bx}{n}

\title{Learning De-Biased Representations for Remote-Sensing Imagery}

\author{%
  Zichen Tian \quad
  Zhaozheng Chen \quad
  Qianru Sun\\
  School of Computing and Information Systems \\
  Singapore Management University \\
  \texttt{\{zichen.tian.2023,zzchen.2019\}@phdcs.smu.edu.sg, qianrusun@smu.edu.sg}
}

\begin{document}

\maketitle

\begin{abstract}
Remote sensing (RS) imagery, requiring specialized satellites to collect and being difficult to annotate, suffers from data scarcity and class imbalance in certain spectrums. Due to data scarcity, training any large-scale RS models from scratch is unrealistic, and the alternative is to transfer pre-trained models by fine-tuning or a more data-efficient method LoRA~\citep{hu2021lora}. Due to class imbalance, transferred models exhibit strong bias, where features of the major class dominate over those of the minor class. In this paper, we propose \texttt{deb}LoRA---a generic training approach that works with any LoRA variants to yield \texttt{deb}iased features. It is an unsupervised learning approach that can diversify minor class features based on the shared attributes with major classes, where the attributes are obtained by a simple step of clustering.
To evaluate it, we conduct extensive experiments in two transfer learning scenarios in the RS domain: from natural to optical RS images, and from optical RS to multi-spectrum RS images. %
We perform object classification and oriented object detection tasks on the optical RS dataset DOTA and the SAR dataset FUSRS.
Results show that our \texttt{deb}LoRA consistently surpasses prior arts across these RS adaptation settings, yielding up to 3.3 and 4.7 percentage points gains on the tail classes for natural $\to$ optical RS and optical RS $\to$ multi-spectrum RS adaptations, respectively, while preserving the performance on head classes, substantiating its efficacy and adaptability~\footnote{~Code: \href{https://github.com/doem97/deblora}{https://github.com/doem97/deblora}}.
\end{abstract}

\vspace{-.8em}
\section{Introduction}\label{sec:intro}
\vspace{-.5em}

Remote sensing (RS) is crucial in various applications such as environmental monitoring, resource management, and disaster response~\citep{zhu2017deep,ma2019deep}.
RS data is collected by various sensors and has multiple spectrums, including optical RS imagery (dubbed as ORS, 400--700nm)~\citep{li2020object}, multi-spectral RS imagery (MSRS, 400--2500nm)~\citep{cong2022satmae}, and synthetic aperture radar imagery (SAR, 1mm-1m)~\citep{shaw2003spectral,sarhandbook}.
These spectrums differ significantly in imaging mechanisms, leading to distinct data characteristics and processing pipelines~\citep{zhu2021deep}.
Given this diversity, learning robust and generic representation models for such data is desirable to reduce processing costs and complexities.

Recently, in natural image domains, large-scale pre-trained visual foundation models (\textit{e.g.}, CLIP~\citep{clip}, Stable Diffusion~\citep{Rombach2021HighResolutionIS}, and DINO~\citep{caron2021emerging}) have shown great advances in robustness and generalization ability.
The zero-shot features extracted from the models show impressive performance in downstream tasks such as object classification, detection and semantic segmentation~\citep{zhang2024vision}, even outperforming the supervised models trained on the specific datasets of those tasks.
However, in the RS domain, training such foundation models from scratch remains challenging.
Even though some trials have been made in past years~\citep{cong2022satmae, guo2023skysense}, their works have clear limitations.
First, they require large-scale RS data for effective training, which are available for only ORS but not other spectrums such as SAR and MSRS~\citep{paolo2022xview3, ding2021object, Guo2022SARID}.
Collecting and annotating images in ``other'' spectrums is difficult due to many factors such as military restrictions, sensor availability, and high acquisition costs, so the data scarcity is unlikely to be alleviated in the near future~\citep{zhu2021deep}.
Second, their works are constrained in small- or medium-scale models, \textit{i.e.}, they use ViT-L (300M) in~\citep{cong2022satmae} and Swin-L (197M) in~\citep{guo2023skysense}, while the foundation models in the natural image domain are much larger (\textit{e.g.}, Latent Diffusion has 860M, and OpenCLIP-H/14 has 986M).
Third, their training-from-scratch approaches are computationally inefficient, requiring a huge amount of GPU memory (VRAM). For instance,~\citep{guo2023skysense} reported the need of 80 * A100 GPU with 80GB VRAM each, totaling 6.4TB.

Instead of learning a foundation model from scratch, we propose to transfer existing foundation models to RS domains. This approach is both data-efficient and 
computation-efficient.
We answer two questions: 1) Which foundation models to transfer? 2) Which transfer learning methods to use?

\begin{wrapfigure}{r}{0.52\textwidth}
  \centering
  \vspace{-1.2em}
  \includegraphics[width=0.51\textwidth]{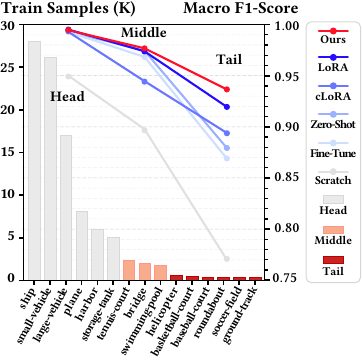}
  \vspace{-.6em}
  \caption{
    \textbf{Long-tailed Problems.} This figure shows 
    1)~ORS datasets (take DOTA~\citep{xia2018dota} as an example) have the long-tailed distribution issue. 
    2)~Model adaptation methods suffer from weak performance in tail classes. 
  }\label{fig:teaser_a}
  \vspace{-.7em}
\end{wrapfigure}
For the first question, we consider foundation models pre-trained on natural images (\emph{e.g.}, CLIP~\citep{clip}, Stable Diffusion~\cite{Rombach2021HighResolutionIS}) as well as the models from remote sensing (RS) images (\emph{e.g.}, SkySense~\citep{guo2023skysense}).
A positive aspect of these models is that they contain the semantic knowledge necessary for learning a new RS domain. However, a great challenge is the large domain gap between natural images and RS domains, or between different RS spectrums. In our preliminary study, we conduct validation experiments.
Fortunately, we observe successful transfer results both from natural to ORS in Figure~\ref{fig:teaser_a} and between different RS spectrums in Table~\ref{tab:main_comp}, 
when compared to the method of TRS-Res101~\citep{zhang2021trs} which does not perform any transfer learning.
The success of natural$\rightarrow$ORS is due to the shared underlying visual elements like edges, textures, and contours, which are intrinsic to both natural and RS images.
The success of ORS$\rightarrow$other RS is 
due to the shared spatial structures, \textit{e.g.}, urban areas, buildings, and object outlines, in different RS spectrums. 

\begin{figure*}[t]
    \centering
    \vspace{-.5em}
    \includegraphics[width=0.9\linewidth]{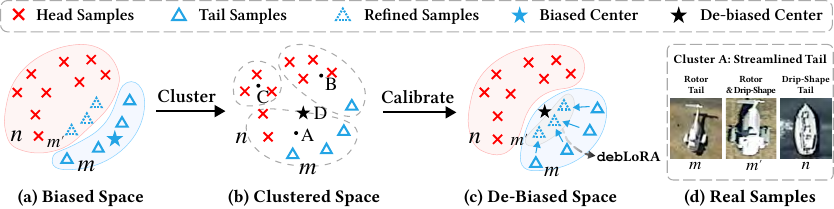}
    \vspace{-.5em}
    \caption{
            \textbf{Two key steps of \texttt{deb}LoRA: feature clustering and calibration.}
            (a) The baseline LoRA feature space is biased towards head classes. Red crosses \includegraphics[height=.7em]{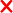} represent head class samples, and blue triangles \includegraphics[height=.7em]{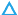} represent tail class samples. The blue star \includegraphics[height=.9em]{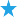} indicates the center of tail class samples. Dashed blue triangles \includegraphics[height=.7em]{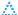} show the validation samples of the tail class wrongly embedded in the head class region, indicating the model bias towards head classes.
            (b) We cluster all features (clusters denoted by gray dotted boundaries) regardless of class labels. $A$, $B$ and $C$ are cluster centers used to generate a de-biased center $D$, as in Eq.~2.
            (c) We calibrate the tail class features by ``moving'' them closer to $D$, as in Eq.~3.
            After these steps, we train the \texttt{deb}LoRA module on the calibrated features of tail classes (together with the original head class features).
          }\label{fig:teaser_b}
    \vspace{-1.5em}
\end{figure*}

For the second question, we found that data-efficient transfer learning methods on foundation models exhibit a strong bias towards major classes.
As shown in Fig.~\ref{fig:teaser_a}, both Fine-Tune and LoRA have significantly lower F1 scores for tail classes.
This is because their learned feature space is biased towards the discriminative features of head classes while neglecting the tail~\citep{yang2020rethinking}.
Taking the head class \texttt{ship} (which takes 28.35\%) and tail class \texttt{helicopter} (0.64\%) as examples on the DOTA dataset~\citep{xia2018dota}.
Fig.~\ref{fig:teaser_b}(a) shows biased LoRA features of ``oval tail'' in the \texttt{ship} sample $n$ and ``rotor tail'' in the \texttt{helicopter} sample $m$. 
We say biased because the LoRA fails to understand the ``oval tail with a rotor'' in another \texttt{helicopter} sample $m'$ and embeds $m'$ wrongly as a \texttt{ship} sample in the feature space.
Please note that the real feature distribution is shown in Figure~\ref{fig:cluster_dist} to support the illustration of Figure~\ref{fig:teaser_b}.
This long-tail issue is particularly severe for transfer learning in the RS domain due to two reasons.
\emph{First, RS datasets suffer from more severe data imbalance than natural image datasets.}
For instance, the imbalance ratios\footnote{~The imbalance ratio is measured by $n_1/n_k$, where $1$ and $k$ are the largest and smallest categories. It reflects the severity of data imbalance~\citep{zhang2023deep}.}
of RS datasets DOTA and ShipRSImageNet reach 86 and 112, respectively, while CIFAR100-LT~\citep{cao2019learning}, a natural image dataset with a similar data scale, has a ratio of only 50. 
This is because annotating under-represented tail class samples in RS, \textit{e.g.}, identifying a rare naval vessel, such as the ``Nimitz'', from SAR image, requires a high level of domain expertise.
\emph{Second, the data scarcity in RS domains determines that RS adaptation methods must be data-efficient}, such as LoRA. 
However, as shown in Table~\ref{tab:ranks}, using fewer parameters in LoRA (being more data-efficient) exacerbates long-tail issues. 
The reason is that this restricts the model capacity and forces the model to prioritize a limited number of features---usually from head classes.

To mitigate this bias without needing more data or labels
in tail classes, we propose an unsupervised learning approach, \texttt{deb}iased LoRA, dubbed \texttt{deb}LoRA.
\texttt{deb}LoRA is based on the features extracted from LoRA (or a LoRA variant) and is generic to LoRA variants.
To be concise, we use LoRA in the following to represent itself and its variants.
Given the LoRA features, \texttt{deb}LoRA has three steps: clustering, calibration, and training. 
First, it clusters all the features regardless of class labels by $K$-means. Each obtained cluster center represents an attribute from one or shared by multiple classes.
Second, these cluster centers are used to calibrate the LoRA features of tail classes and enhance the territory of tail classes in the feature space. 
We illustrate these two steps in Figure~\ref{fig:teaser_b}.
Last, the calibrated features are used as the learning objectives to train a \texttt{deb}LoRA module with a similar network architecture to LoRA. 
The learned \texttt{deb}LoRA is thus a de-biased feature extractor.

We observe that after $K$-means clustering, each cluster center captures a general visual attribute shared across different classes. For instance, in Figure~\ref{fig:teaser_b}(b), cluster $A$ corresponds to the general vehicle attribute ``streamlined tail'', which includes both head class sample $n$ and tail class sample $m$. Such clusters can thus yield a balanced representation base, making the tail more robust by integrating common attributes with the head. 

One may ask ``what if some attributes are dominated by the attribute features of head classes?''
We address this question by proposing a weighting scheme, in the step of calibration. 
In specific, for each tail class sample (\textit{e.g.}, $m$ in Fig.~\ref{fig:teaser_b}(c)), we calibrate it by forcing its feature closer to the de-biased center ($D$)---the weighted average of all cluster centers.
The weights are determined by the number of samples in each cluster, ensuring that this center is not dominated by clusters with mostly head class samples.
This calibration process results in de-biased representations that capture a more comprehensive range of visual attributes shared across classes, leading to improved features of tail classes (\emph{e.g.}, $m'$).
Lastly, we re-train a LoRA module to map biased representations towards these debiased centers.
Please find more details of justifications in Sec.~\ref{sec_justification}.
Our method significantly improves the features of tail classes. Moreover, it is efficient as it learns only a lightweight low-rank module while keeping the original foundation model frozen. 

Our contributions can be concluded three-fold: 1) We demonstrate the effectiveness of adapting foundation models for data-scarce RS domains. 2) We propose Incremental LoRA, a novel method that de-biases category-specific representations for long-tailed RS adaptation. 3) We conduct extensive experiments to validate our approach on multiple RS adaptation settings and downstream tasks.

\vspace{-1em}
\section{Related Works}\label{sec:related_works}

\noindent\textbf{Representation Learning for RS Images.}
Self-supervised representation learning in RS image domains mainly includes contrastive- and generative-based methods.
\textbf{Contrastive-based} methods, such as Tile2vec~\citep{jean2019tile2vec}, Seasonal contrast~\citep{manas2021seasonal} and SauMoCo~\citep{kang2020deep}, heavily rely on rich temporal data or high-resolution samples, which are often unavailable for data-scarce RS spectrums~\citep{wang2022self}.
\textbf{Generative-based} methods, such as RR-SSL~\citep{zhang2019rotation} and SGSAGANs~\citep{guo2021self}, reconstruct inputs to capture the global data distribution and learn fine-grained patterns. However, they require large-scale data to form robust latent space~\citep{gui2021review}.
Recently, \textbf{foundation models} in the RS domain, such as SatMAE~\citep{cong2022satmae}, SpectralGPT~\citep{hong2023spectralgpt}, and SkySense~\citep{guo2023skysense}, have shown promising results for ORS tasks. 
SpectralGPT~\citep{hong2023spectralgpt} tackles spectrum diversity by pre-training separate tokenizers for each spectrum, which still needs large amounts of data.
Another problem is that existing RS foundation models are much smaller than those in the natural image domain (\textit{e.g.}, SatMAE-L~\citep{cong2022satmae} has 300M parameters \textit{v.s.} 986M of OpenCLIP-H/14~\citep{cherti2023reproducible}).
Instead of learning RS foundation models from scratch, we propose to adapt them from pre-trained models to RS. Our approach 1) greatly reduces the computational cost, 2) can be easily adapted to various data-scarce RS spectrums, and 3) benefit from the strong representation power of large-scale foundation models in other domains.

\noindent\textbf{Long-tailed Data Distribution and its Bias Problem.} 
Long-tailed data distribution, where a few head classes cover most of the samples, is prevalent in both natural and RS image domains~\citep{van2017devil,zhang2023deep}.
This imbalance leads to biased feature representations, where the model focuses on discriminative features for head classes while neglecting subtle but crucial features for tail classes~\citep{zhang2023deep,kang2020exploring}.
Zhang et al.~\citep{zhang2023deep} observed that such a feature space is usually broader for head classes than tail classes, and the decision boundary tends to be biased towards head classes, \textit{i.e.}, many false positive predictions for head classes.
Existing solutions include sample-level, meta-learning, and representation-level approaches~\citep{zhang2023deep}: 
\textbf{Sample-level} methods, such as re-sampling~\citep{shen2016relay} and data augmentation~\citep{chu2020feature}, aim to directly balance the sample distribution.
However, they require sample annotations~\citep{cao2019learning,shen2016relay} or rely on data diversity~\citep{chu2020feature}, both of which are unrealistic in the data-scarce RS spectrums such as SAR~\citep{sarhandbook} and MSRS~\citep{cong2022satmae}.
\textbf{Meta-learning} methods~\citep{jamal2020rethinking,wang2017learning} formulate the problem as ``learning to learn'' and adapt the model to a balanced meta-test set.
They depend on the data diversity of the training sets and the availability of balanced validation sets, and therefore, are less applicable for data-scarce RS domains.
The \textbf{representation-level} methods enhance the learned representation space, including metric learning losses~\citep{huang2016learning}, margin-based losses~\citep{cao2019learning}, and feature transfer from head to tail classes~\citep{liu2020deep,yin2019feature}.
However, they are designed for supervised single-domain settings and do not address the challenges of model adaptation to RS:
1) handling multiple downstream tasks (\textit{e.g.}, small object detection, scene segmentation, change detection),
and 2) multiple spectrums (such as ORS and SAR).
In contrast, we propose an \textit{unsupervised adaptation} method to tackle these challenges in this paper.

\noindent\textbf{Transfer Learning in Remote Sensing.}
Transfer learning in remote sensing primarily focuses on adaptation within the optical imagery domain.
They can be categorized into supervised and unsupervised methods.
Supervised methods \citep{fernando2013unsupervised,long2013transfer,rajan2008active,persello2012active,matasci2012svm} align distributions using target labels. 
However, they require task-specific annotations, which are scarce in SAR and multispectral domains and limit the applicability of the obtained models to multiple downstream tasks.
Unsupervised DA (UDA) methods aim to learn domain-invariant features without requiring labeled data in the target domain, including transfer component analysis~\citep{pan2010domain,matasci2015semisupervised}, manifold alignment~\citep{tuia2014semisupervised,yang2015domain,yang2015spectral}, and adversarial learning~\citep{bejiga2019domain,elshamli2017domain,tasar2020standardgan}. 
However, they are designed for single-source, single-target adaptation within the same spectrum~\citep{pacifici20112011,martini2021domain}.
Besides, the manifold alignment and adversarial methods require significant computational resources, often involving the training of several copies of the source model, while component analysis methods involve complex pipelines.
These factors make them unsuitable for foundation models, which are already computationally intensive.
In contrast, our method tackles multi-spectrum adaptation without requiring extra labels.
It is also computationally efficient.
\vspace{-0.5em}
\section{LoRA and cLoRA}
\vspace{-0.5em}
Our \texttt{deb}LoRA is based on the LoRA~\cite{hu2021lora} or its variants~\citep{yue2024few}, but is orthogonal and generic to them. 

\noindent\textbf{LoRA.} LoRA was initially proposed to adapt a pre-trained large-scale language model to downstream tasks. 
It assumes adapted parameters are sparse during model training when the data is limited.
It introduces a low-rank factorization of the difference between original and adapted parameters, \textit{i.e.}, $\Delta \theta:=B{\cdot} A$.
Here, $\theta \in \mathbb{R}^{d\times k}$ represents the parameters of pre-trained model, and $B\in \mathbb{R}^{d\times r}$ and $A\in \mathbb{R}^{r\times k}$ denote low-rank factors, with $r\ll \min(d,k)$. The updated parameters $\hat{\theta}$ are thus given by $\hat{\theta}=\theta {+} \Delta \theta = \theta {+} B{\cdot} A$. 
During inference, the obtained LoRA modules could be combined through a weighted sum,
$\hat{\theta} = \theta + \sum_{i} w_i \Delta \theta_{i}$, where $w_i$ denotes combination weights.

\noindent\textbf{cLoRA.} 
To tackle the long-tailed issue of LoRA, we also explore its variant cLoRA~\citep{yue2024few}.
The key idea of cLoRA is to learn a separate LoRA module for each class, denoted as $\Delta \theta_c$ for class $c$, to ensure that the learned representations of one class do not interfere with those of other classes.
Formally, the adapted parameters for class $c$ are given by $\hat{\theta}_c = \theta + \Delta \theta_c = \theta + B_c \cdot A_c$, where $B_c \in \mathbb{R}^{d \times r}$ and $A_c \in \mathbb{R}^{r \times k}$ are the low-rank factors specific to class $c$.
During training, each cLoRA module $\Delta \theta_c$ is optimized using only the data from class $c$, allowing it to capture class-specific features.
During inference, as there is no class label available, we use all the cLoRA modules to extract features for the input.
Specifically, for an input $x$, we obtain the features $z_c = \hat{\theta}_c(x)$ using each cLoRA module $\hat{\theta}_c$.
The final feature representation is then obtained by concatenating the features from all the cLoRA: $z=[z_1; z_2; \ldots; z_C]$, where $C$ is the total number of classes.

\section{De-biased LoRA (\texttt{deb}LoRA)}\label{sec_approach}
The algorithm of \texttt{deb}LoRA consists of two steps: generating debiased features, and then using them to train a \texttt{deb}LoRA module.
In the first step, we perform unsupervised clustering on biased feature space $\mathcal{Z}$ (\textit{i.e.}, composed by original LoRA features biased to head classes) to obtain debiased features $\hat{\mathcal{Z}}$.
In the second step, we use $\hat{\mathcal{Z}}$ as the learning target to train a \texttt{deb}LoRA module. 
The \texttt{deb}LoRA learns the mapping between biased and de-biased features.
We justify the feasibility of learning such a mapping in Section~\ref{sec_justification}.

\subsection{Problem Formulation}
Given a pre-trained 
feature extractor
$f: \mathcal{X} \rightarrow \mathcal{Z}$ and a long-tailed RS dataset $\mathcal{D} = {(x, y)}$, where $x \in \mathcal{X}$ is an RS image, $y \in \mathcal{Y}$ is its annotation and $\mathcal{Z}$ is the biased feature space\footnote{~We define feature space $\mathcal{Z}$ as biased if $\text{Vol}(\mathcal{Z}_h) \gg \text{Vol}(\mathcal{Z}_t)$, and $\exists z_t \in \mathcal{Z}_t : P(z_t \in \mathcal{Z}_h) > P(z_t \in \mathcal{Z}_t)$, where $\mathcal{Z}_h$ and $\mathcal{Z}_t$ denotes the feature spaces of head and tail classes respectively, $\text{Vol}(\cdot)$ denotes feature space volume, and $P(\cdot)$ denotes the probability predicted by the model.}, our goal is to adapt $f$ to the target dataset $\mathcal{D}$ while yielding a de-biased feature space $\hat{\mathcal{Z}}$, 
\textit{i.e.}, adapted encoder is $\hat{f}: \mathcal{X} \rightarrow \hat{\mathcal{Z}}$.
The de-biased feature representation $\hat{\mathcal{Z}}$ should improve downstream task performance on tail classes without sacrificing the performance on head classes.

\subsection{Stage 1: Representation De-biasing}
\paragraph{Feature Clustering.}
Given a pre-trained encoder $f_\theta: \mathcal{X} \rightarrow \mathcal{Z}$ that maps input images to a biased representation space, where $f_\theta$ is parameterized by $\theta$, we first extract features for each sample in the dataset: $z_i = f_{\theta}(x_i)$, $i\in N$. 
We then apply $K$-means clustering on $\{z_i\}$ to obtain $K$ clusters. To mitigate imbalanced clusters, we impose a constraint that each cluster should contain at least $\frac{N}{K\cdot\rho}$ samples, where $\rho$ is a pre-defined constant. 
The clustering objective is:
\begin{equation}
\label{eq:kmeans}
\min_{\mu_k} \sum_{i=1}^N \min_{k} \|z_i - \mu_k\|^2, \quad \text{s.t.} \;\; \forall k, \; n_k \geq \frac{N}{K\cdot\rho},
\end{equation}
where $\mu_k$ and $n_k$ denote the center and size of the $k$-th cluster, respectively.

\paragraph{De-biased Cluster Centers.} 
For each tail class $c$, we calculate its de-biased representation center $\hat{\mu}_c$ by weighted averaging all the cluster centers:
\begin{equation}
\label{eq:center}
\hat{\mu}_c = \sum_k w_k \cdot \mu_k, \quad \text{where} \;\; w_k = \frac{n_{k}}{n_c}.
\end{equation}
Here $n_k$ denotes the number of samples from class $c$ in the $k$-th cluster, and $n_c$ is the total number of samples in class $c$. 
The weight $w_k$ is proportional to the fraction of class $c$ samples in the $k$-th cluster. 
This ensures that the de-biased center $\hat{\mu}$ is not dominated by head classes.

\subsection{Stage 2: De-Biased Low Rank Adaptation (\texttt{deb}LoRA)}\label{sec:stage2}

\paragraph{Tail Class Calibration.}
For each tail class sample $x$ with representation $z$, we calibrate $z$ by moving it closer to the de-biased center $\hat{\mu}$:
\begin{equation}
\label{eq:calibrate}
\tilde{z} = \alpha z + (1 - \alpha) \hat{\mu},
\end{equation}
where $\alpha \in [0, 1]$ is a hyper-parameter controlling the degree of calibration.
We empirically set $\alpha$ based on the imbalance ratio $\gamma$ of each tail class: $\alpha = \min(1, \frac{10}{\gamma})$.
For tail classes with larger imbalance ratio, a higher $\alpha$ encourages the calibrated representation $\tilde{z}$ to be closer to the de-biased center $\hat{\mu}$, as the original representation $z$ is less reliable due to its learning from limited samples.
While for classes with smaller $\gamma$, a lower $\alpha$ is used to retain the discriminative information of $z$.
For instance, the DOTA dataset's tail class \texttt{helicopter} has high $\gamma=45.45$, so its $\alpha$ reaches $0.22$.

\paragraph{Learning \texttt{deb}LoRA.}
With the pre-trained encoder $f_{\theta}$ frozen, we learn a LoRA module $g_\phi: \mathcal{Z} \rightarrow \hat{\mathcal{Z}}$ parameterized by $\phi$ to map the biased representations to the calibrated ones. The training objective is:
\begin{equation}
\label{eq:learn}
\min_\phi \frac{1}{|\mathcal{D}_t|} \sum_{x \in \mathcal{D}_t} \|g_\phi(f_{\theta}(x)) - \tilde{z}\|^2,
\end{equation}
where $\mathcal{D}_t$ is the set of tail class samples. During inference, we apply the learned LoRA module to extract the de-biased representations $z = g_\phi(f_{\theta}(x))$
for an input image $x$.
The complete algorithm of \texttt{deb}LoRA is summarized in Algorithm~\ref{alg:unblora}.
\vspace{-1em}

\begin{algorithm*}[h!] 
\caption{\texttt{deb}LoRA}\label{alg:unblora}
\begin{algorithmic}[1]
\Require%
Long-tailed training set $\mathcal{D} = \{(x, y)\}$, pre-trained encoder $f_{\theta}: \mathcal{X} \rightarrow \mathcal{Z}$, number of clusters $K$, balance factor $\rho$
\Ensure%
A LoRA module $g_\phi$ that de-biases $f_\theta$
\State%
Extract biased representations $z = f_{\theta}(x)$ for each sample $x \in \mathcal{D}$ using pre-trained $f_{\theta}$ 
\State%
Perform constrained $K$-means clustering on $\{z\}$ (\eqref{eq:kmeans}) to obtain cluster centers $\{\mu_k\}_{k=1}^{K}$, where each cluster has at least $\frac{N}{K\cdot\rho}$ samples  
\For{each tail class $c$}
    \State%
    Calculate its de-biased representation center $\hat{\mu}_c$ by weighted averaging all cluster centers $\{\mu_k\}_{k=1}^{K}$ (\eqref{eq:center})
    \For{each sample $x \in \mathcal{D}_c$}
        \State%
        Extract biased representation $z = f_{\theta}(x)$
        \State%
        Calibrate $z$ to $\tilde{z}$ by moving it closer to $\hat{\mu}_c$ with factor $\alpha = 10/\gamma$ (\eqref{eq:calibrate})
    \EndFor%
\EndFor%
\State%
Learn a LoRA module $g_\phi: \mathcal{Z} \rightarrow \hat{\mathcal{Z}}$ to map biased representations to calibrated ones
\State%
\Return%
$g_\phi$
\end{algorithmic}
\end{algorithm*}

\begin{figure}[t]
    \centering
    \vspace{-1em}
    \includegraphics[width=\linewidth]{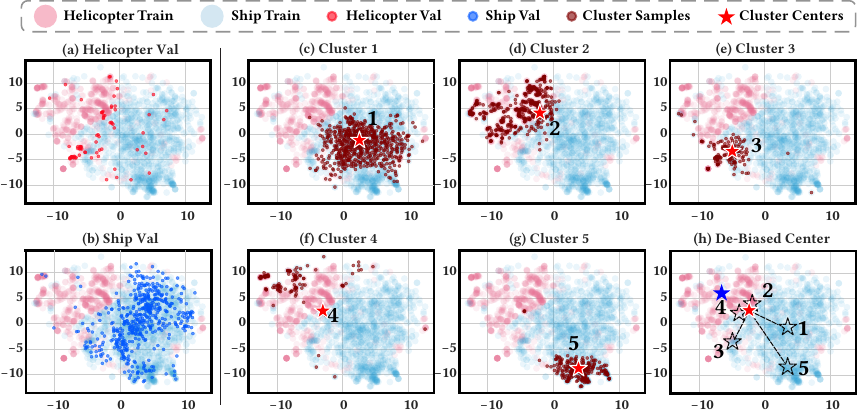}
    \vspace{-1.5em}
    \caption{
    \textbf{t-SNE visualization of validation samples and clusters.}
    The first column shows the distribution of \texttt{helicopter} (tail) and \texttt{ship} (head) validation samples.
    Subfigures (c)-(g) are the clusters and their centers when $K$=5 in $K$-means.
    In (h), the dotted lines and stars indicate that we compute a de-biased center for the tail class (\texttt{helicopter}) by weighted averaging the five cluster centers, and the blue star is the original biased center of \texttt{helicopter} training samples.
    }\label{fig:cluster_dist}
    \vspace{-1em}
\end{figure}

\vspace{-1em}
\subsection{Justification}\label{sec_justification}
\vspace{-1em}
We discuss the biased representation space of LoRA, and then justify the effectiveness of our three critical operations in \texttt{deb}LoRA: \textbf{clustering}, \textbf{weighting}, and \textbf{calibration}. 
We show the real sample distribution in Figure~\ref{fig:cluster_dist} and an illustrative example in Figure~\ref{fig:teaser_b}.

\vspace{-1em}
\paragraph{LoRA is Biased.}
The feature space learned by LoRA is biased towards head classes~\citep{yang2020rethinking}, evidenced by two observations. 
1)~The head class representations over-expand their territory into the tail class space. As shown in Figure~\ref{fig:cluster_dist}, most of the \texttt{ship} (head) validation samples are distributed within its own representation space, while many \texttt{helicopter} (tail) validation samples are wrongly distributed in the \texttt{ship}'s space. 
2) The center of the entire space is biased towards head class, as the \texttt{ship} training samples significantly overlap with the \texttt{helicopter} training samples.
This bias occurs because, during training, the encoder is exposed to significantly more diverse samples of head class. 
\vspace{-.8em}

\paragraph{Clustering.}
By feature clustering, we obtain a set of cluster centers that benefit the tail classes in two ways.
1) \textit{Improved robustness.} 
The obtained cluster centers, shown as red stars in Figure~\ref{fig:cluster_dist}(c)-(g), represent visual prototypes\citep{caron2018deep}, \textit{i.e.}, general visual attributes common to both head and tail classes, such as ``streamlined tail'' or ``with wooden deck''. 
These cluster centers are more robust than the original tail class representations because they leverage the diversity of head class samples.
2) \textit{Reduced imbalance.}
Certain clusters exhibit reduced long-tail issues.
The clusters in Figure~\ref{fig:cluster_dist}(d)-(f) contain more samples from \texttt{helicopter} than \texttt{ship}. 
This is because the clusters are formed based on intrinsic visual similarities among images, regardless of their imbalanced class labels.
Using these cluster centers avoids the risk of tail class attributes (\textit{e.g.}, ``rotor tail'' and its variants in \texttt{helicopter}) being overwhelmed by head class attributes (\textit{e.g.}, ``oval tail'' and its variants in \texttt{ship}).

\vspace{-.8em}

\paragraph{Weighting and Calibration.}
One might ask, ``Are the data imbalances within each cluster or among different clusters still issues?''
\textit{E.g.}, the 5-th cluster in Figure~\ref{fig:cluster_dist} contains only \texttt{ship} samples and seems irrelevant to \texttt{helicopter}.
To answer this, we perform the \textit{weighted averaging} over cluster centers, and the \textit{calibration} over tail class samples:
1) \textit{Weighted averaging.}
When calculating the de-biased representation center for each tail class (\eqref{eq:center}), we assign higher weights to clusters containing a larger fraction of that particular tail class.
The de-biased center (red star in Figure~\ref{fig:cluster_dist}(h)) better captures the true distribution of the validation samples of \texttt{helicopter}, compared to the original biased center (blue star in Figure~\ref{fig:cluster_dist}(h)).
2) \textit{Calibration.}
We calibrate the representation of each tail class sample by moving it closer to the class's de-biased center (\eqref{eq:calibrate}).
The calibration factor $\alpha$ is inversely proportional to the imbalance ratio of the tail class.
This design ensures severely underrepresented tail classes like \texttt{helicopter} receive stronger calibration.
\vspace{-.5em}
\section{Experiments and Analyses}\label{sec:exp}
\vspace{-.7em}

We evaluate our \texttt{deb}LoRA on two settings: 1)~adapting natural image foundation models to RS, and 2)~adapting ORS foundation models to SAR. 
For the first setting, we conduct experiments on two representative RS tasks: object classification and oriented object detection.
For the second setting, we conduct experiments on a representative SAR task---fine-grained ship classification.

\noindent\textbf{Natural $\to$ RS adaptation.}
1)~\emph{Foundation model.} 
We use two state-of-the-art foundation models: Stable Diffusion v1.5 (SD)~\citep{Rombach2021HighResolutionIS} and OpenCLIP~\citep{ilharco_gabriel_2021_5143773}. 
Both models have shown impressive generalization ability on various tasks when adapted to domains like medical images~\citep{Wilde2023MedicalDO}. 
However, their transferability from natural images to the RS domain remains under-explored.
2)~\textit{RS dataset.}
We use the DOTA dataset~\citep{ding2021object}, a large-scale benchmark for RS object recognition.
DOTA contains 188,282 instances from 15 categories, covering various scales, orientations, and shapes.
We define the long-tail split as follows: 6 classes with <1\% instances as tail, 3 classes with 1\%-5\% instances as middle, and the remaining 6 classes (each with >5\% instances) as head.
This split exhibits a clear long-tail distribution, evidenced by the performance gap between head and tail classes for the baseline methods (see Table~\ref{tab:ablation} row 1).
3)~\textit{Tasks.}
For the classification task, we obtain features from the adapted foundation models and train a linear classifier. We report the macro F1-score that fairly evaluate the performance across all classes.
For detection, we train a FCOS detector head~\citep{tian2020fcos} on obtained representations and evaluate using the mAP.

\noindent\textbf{ORS $\to$ SAR adaptation.}
1)~\textit{Foundation model.}
We use SatMAE-L~\citep{cong2022satmae}, the state-of-the-art open-sourced foundation model for RS. 
SatMAE-L is pre-trained on large ORS datasets using self-supervised learning. It has $307$M parameters and requires 6,144 GPU hours to train from scratch.
2)~\textit{SAR dataset.}
We evaluate our method on the fine-grained ship classification task of SAR.
Existing SAR ship datasets have insufficient samples to evaluate the model performance reliably, \textit{e.g.}, only 2 samples in test set for tail class ``WingInGrnd'' on the FUSAR-Ship dataset.
We thus create a new dataset by combining two high-resolution ($<$10m/pixel) SAR datasets: FUSAR-Ship~\citep{hou2020fusar} and SRSDD~\citep{lei2021srsdd}. Details of this combined dataset are provided in the Appendix.
3)~\textit{Ship classification task.}
We follow the same setup as in the natural $\to$ RS setting for this SAR task.

\noindent\textbf{Implementation Details.}
1)~\textit{Fine-tuning baseline.}
We fine-tune the foundation models until the training loss stabilizes.
During inference, we use null prompts as no ground truth is available.
For SD, we extract features from the U-Net after applying one denoising step~\citep{tang2023emergent}.
For OpenCLIP, we extract features from its visual encoder's final layer before the projection head.
2)~\textit{LoRA and variants.}
We apply LoRA modules to all linear layers in the foundation models.
We use a rank of 8 for LoRA, as it suffers from the most severe long-tail issues. We also validate our method with higher ranks (\textit{e.g.}, 64) in Table~\ref{tab:ranks}.
During inference, we extract features from the U-Net encoder output followed by global average pooling (GAP).
For cLoRA, we concatenate the category-specific features after GAP.
3)~\textit{\texttt{deb}LoRA.}
The \texttt{deb}LoRA involves two hyperparameters: the calibration factor $\alpha$, and the number of clusters $K$.
We set $\alpha$ as inversely proportional to the imbalance ratio of the tail class, as described in Section~\ref{sec_justification}.
We empirically set $K$=32 (ablation study on $K$ are provided in Appendix).

\noindent\textbf{Evaluation Metrics.}
1)~\textit{Classification.}
We use linear probing (i.e., train a linear classifier on the top of frozen features) to evaluate the learned representations~\citep{he2020momentum,he2022masked}.
It is simple and avoids introducing additional learning operations.
We apply GAP and ReLU on the extracted features before linear probing.
We report the macro F1-score, which assigns equal importance to all classes---more suitable for evaluating imbalanced datasets.
We report scores for head, middle, and tail classes separately, as well as the overall score averaged across all categories.
2)~\textit{Detection.}
We use the lightweight FCOS~\citep{tian2020fcos}, an anchor-free detector head, to avoid potential interference from pre-defined anchors.
We extract high-resolution feature maps from the SD U-Net output.
During feature clustering and re-training, we use per-instance features for each category.
During inference, we extract features from the entire image and feed them to the detector head.
We report the mAP metric.

\begin{wraptable}{r}{0.5\textwidth}
\centering
\vspace{-6.5mm}
\caption{\textbf{Ablation study of \texttt{deb}LoRA.} 
We apply our \texttt{deb}LoRA based on LoRA and cLoRA.
Results are reported for the adaptation from SD $\to$ DOTA recognizer.
Params (M) refers to the number of updated parameters during the adaptation.
Our results are marked in \setlength{\fboxsep}{2.5pt}\colorbox{gray!20}{gray}.
}\label{tab:ablation}
\resizebox{0.5\textwidth}{!}{%
\begin{tabular}{lccccc}
\toprule[1.3pt]
\multirow{2}{*}{\textbf{Method}} & \multicolumn{4}{c}{\textbf{Macro F1 Score (\%)}} & \multirow{2}{*}{\textbf{Params (M)}} \\ \cmidrule(lr){2-5}
& Head & Middle & Tail & Overall & \\
\midrule \addlinespace[0.05em] \midrule
Zero-Shot & 99.2 & 97.3 & 87.8 & 94.3 & --- \\
Fine-Tune & 99.1 & 96.7 & 86.8 & 93.7 & 860 \\ \midrule
cLoRA & 99.1 & 94.3 & 89.3 & 94.2 & 0.08 \\
\rowcolor{gray!15} \; w/ \texttt{deb}LoRA & \textbf{99.3} & \textbf{97.5} & \textbf{93.5} & \textbf{96.6} & 0.08 \\ \midrule
LoRA & \textbf{99.4} & 97.2 & 91.8 & 95.9 & 0.08 \\
\rowcolor{gray!20} \; w/ \texttt{deb}LoRA & 99.1 & \textbf{98.7} & \textbf{94.5} & \textbf{97.1} & 0.08 \\
\bottomrule[1.3pt]
\end{tabular}
}
\end{wraptable}%
\noindent\textbf{Ablation study.} 
In Table~\ref{tab:ablation}, rows 1 and 2 show the results of using zero-shot features of SD or fine-tuned SD features on DOTA to train RS object recognizers.
Recognizers' performances are strongly biased to head classes---around 12 percentage points drop for tail classes.
From rows 3 and 5, we can see such issues get resolved a bit when using LoRA methods.
Rows 4 and 6 show that \texttt{deb}LoRA significantly outperforms LoRA methods on tail classes---by 4.2 points and 2.7 points, respectively.
Specifically, compared to cLoRA, \texttt{deb}LoRA does not even sacrifice the performance for head classes. 
To quantitatively validate its working mechanism, we analyzed feature discrimination. Results show that \texttt{deb}LoRA enlarges inter-class distances and reduces intra-class distances for tail classes (see Appendix).
In addition, \texttt{deb}LoRA needs just the same amount of parameters as LoRA (0.08M), which is appealing for computation.
\begin{wraptable}{r}{0.5\linewidth}
\centering
\vspace{-1.8em}
\caption{\textbf{Compare LoRA ranks.} 
The table compares different ranks of the LoRA module.
Our results are marked in \setlength{\fboxsep}{2.5pt}\colorbox{gray!20}{gray}.
}\label{tab:ranks}
\resizebox{\linewidth}{!}{%
\begin{tabular}{lccccc}
\toprule[1.3pt]
\multirow{2}{*}{\textbf{Method}}  & \multicolumn{4}{c}{\textbf{Macro F1 Score (\%)}} & \multirow{2}{*}{\textbf{\thead{Params\\(M)}}} \\ \cmidrule(lr){2-5}
& Head & Middle & Tail & Overall &  \\ 
\midrule \addlinespace[0.05em] \midrule
Rank 8  & 99.4 & 97.2 & 91.8 & 96.1 & 0.08 \\
\rowcolor{gray!20} \;\; w/ \texttt{deb}LoRA & 99.1 & 98.7 & 94.5 & 97.1 & 0.08 \\ 
Rank 16  & 99.0 & 95.9 & 92.4 & 95.8 & 0.16 \\
Rank 32  & 99.4 & 96.9 & 93.0 & 96.4 & 0.32 \\
Rank 64  & 99.1 & 96.9 & 94.0 & 96.7 & 0.64 \\
\rowcolor{gray!20} \;\; w/ \texttt{deb}LoRA & 99.1 & 98.7 & 96.2 & 98.0 & 0.64 \\
\bottomrule[1.3pt]
\end{tabular}
}
\vspace{-1em}
\end{wraptable}
\noindent\textbf{LoRA Ranks.}
We investigate the impact of different LoRA ranks on the long-tailed classification performance in Table~\ref{tab:ranks}.
We have two key observations.
1) As the LoRA rank decreases, the performance on tail classes drops more significantly than on head classes. For example, when the rank is reduced from 64 to 8, the F1-score of tail classes decreases by 2.2 percentage points, while that of head classes even increases by 0.3 percentage. 
This supports our hypothesis that the limited parameter capacity of low-rank LoRA forces it to prioritize learning the head classes, exacerbating the long-tail problem.
2) \texttt{deb}LoRA consistently improves the performance on middle and tail classes across different LoRA ranks. Notably, with rank 64, \texttt{deb}LoRA achieves a 2.2 percentage points improvement on tail classes while maintaining the performance on head classes.

\begin{table*}[t]
\centering
\caption{
    \textbf{State-of-the-art comparison under different adaptation settings.}
    The experiments are conducted on two RS adaptation settings: 
    1)~Natural$\to$ORS, where we adopt Stable Diffusion (SD) and OpenCLIP as foundation models and DOTA as the target dataset. 
    2)~ORS$\to$SAR, where we adopt SatMAE as the foundation model and FUSRS (SAR imagery dataset) as the target dataset.
    Results are evaluated by linear probing and reported in macro F1-Score (\%).
    The highest result in each position is highlighted by \textbf{bold}.
    Our results are marked in \setlength{\fboxsep}{2.5pt}\colorbox{gray!20}{gray}.
}\label{tab:main_comp}
\resizebox{\linewidth}{!}{%
\rowcolors{2}{white}{white}
\begin{tabular}{lcccccccccccc}
\toprule[1.3pt]
\multirow{2}{*}{\textbf{Method}} & \multicolumn{3}{c}{\textbf{SD $\to$ DOTA}} & \multicolumn{3}{c}{\textbf{OpenCLIP $\to$ DOTA}} & \multicolumn{2}{c}{\;\; \textbf{SatMAE $\to$ FUSRS}} & \multicolumn{3}{c}{\textbf{Mean}} \\
\cmidrule(lr){2-4} \cmidrule(lr){5-7} \cmidrule(lr){8-9} \cmidrule(lr){10-12}
& Head & Middle & Tail & Head & Middle & Tail & \;\; Head & \;\;Tail & Head & Middle & Tail \\ \midrule \midrule
Zero-Shot & 99.2 & {97.3} & 87.9 & 93.1 & 92.7 & 91.7 & \;\; 78.3 & \;\;67.8 & 90.2 & {95.0} & 82.5 \\
Fine-Tune & 99.1 & 96.7 & 86.8 & 93.1 & 91.1 & 89.2 & \;\; 88.6 & \;\;73.6 & 93.6 & 93.9 & 83.2 \\ \midrule
cLoRA & 99.1 & 94.3 & 89.3 & 97.3 & 93.3 & 92.2 & \;\; 89.9 & \;\;82.0 & 95.5 & 93.8 & 87.9 \\
\rowcolor{gray!15} \quad  w/ \texttt{deb}LoRA & {99.3} & 97.5 & 93.5 & {\textbf{97.6}} & \textbf{95.8} & \textbf{95.0} & \;\; \textbf{92.5} & \;\;\textbf{86.1} & {\textbf{96.5}} & \textbf{96.7} & \textbf{91.5} \\ \midrule
LoRA & 99.4 & 97.2 & 91.8 & 96.6 & 92.7 & 91.6 & \;\; 87.1 & \;\;76.3 & 94.4 & 95.0 & 86.6 \\
\quad  w/ ResLT~\citep{cui2022reslt} & \textbf{99.4} & {97.7} & 93.0 & 97.7 & 94.1 & 93.8 & \;\;86.6 & \;\;75.4 & 94.6 & 95.9 &  87.4\\
\quad  w/ SADE~\citep{zhang2022self} & 99.1 & 97.3 & 92.4 & 97.3 & 93.0 & 92.5 & \;\;89.6 & \;\;78.4 & 95.3 & 95.2 & 87.8 \\
\rowcolor{gray!15} \quad  w/ \texttt{deb}LoRA & 99.3 & \textbf{97.7} & \textbf{95.1} & 97.2 & {95.6} & 94.8 & \;\; 90.1 & \;\;81.0 & 95.5 & \textbf{96.7} & 90.3 \\
\bottomrule[1.3pt]
\end{tabular}
}
\vspace{-.3cm}
\end{table*}
\noindent\textbf{Compare with SOTA.}
1) \textit{Object Classification.}
Table~\ref{tab:main_comp} compares our \texttt{deb}LoRA with state-of-the-art methods under three adaptation tasks.
We draw four key observations from the table.
1)~\texttt{deb}LoRA consistently outperforms LoRA on tail classes across all adaptation tasks, with the largest gain of 4.7 percentage points for ORS $\to$ SAR (\textit{i.e.}, SatMAE $\to$ FUSRS). This shows the consistent efficiency of our approach in tackling the long-tail problem of RS domains.
2)~Compared to SD $\to$ DOTA setting, cLoRA performs exceptionally well under OpenCLIP $\to$ DOTA setting, slightly surpassing LoRA.
We hypothesize that OpenCLIP's feature space aligns particularly well with cLoRA's class-specific
\begin{wraptable}{r}{0.5\linewidth}
\centering
\vspace{-.65em}
\caption{\textbf{Evaluation on the oriented object detection task.} 
We implement \texttt{deb}LoRA for long-tailed detection tasks.
Our results are marked in \setlength{\fboxsep}{2.5pt}\colorbox{gray!20}{gray}.}
\resizebox{\linewidth}{!}{%
\begin{tabular}{lcccc}
\toprule[1.3pt]
\multirow{2}{*}{\textbf{Method}}  & \multicolumn{3}{c}{\textbf{mAP} (\%)$\uparrow$} & \multicolumn{1}{c}{\textbf{Average}} \\ \cmidrule(lr){2-4}
& Head & Middle & Tail & (\%)$\uparrow$ \\ 
\midrule \addlinespace[0.05em] \midrule
Zero-Shot & 71.0 & 73.7 & 55.9 & 66.9 \\
Fine-Tune & 76.3 & 84.9 & 64.3 & 75.2 \\
LoRA & 77.5 & 86.3 & 66.5 & 76.7 \\ 
\rowcolor{gray!16}\;\; w/ Reweight~\citep{kang2019few} & 74.3 & 86.8 & 66.9 & 76.0 \\
\rowcolor{gray!16}\;\; w/ ECM~\citep{hyun2022long} & 78.1 & 87.4 & 68.5 & 78.0 \\
\rowcolor{gray!16}\;\; w/ \texttt{deb}LoRA & \textbf{79.4} & \textbf{88.5} & \textbf{73.2} & \textbf{80.4}\\ 
\bottomrule[1.3pt]
\end{tabular}
}\label{tab:comp_det}
\end{wraptable}
design. However, \texttt{deb}LoRA remains robust across both foundation models.
3)~The performance gains of \texttt{deb}LoRA are most significant for SatMAE $\to$ FUSRS (+4.7 points) compared to SD $\to$ DOTA and OpenCLIP $\to$ DOTA (+3.3 and +3.2 points, respectively). This suggests that our method can leverage domain similarity more effectively when adapting between related image domains (SatMAE and FUSRS are RS datasets).
We think this is because \texttt{deb}LoRA's clustering step captures and utilizes the shared domain-specific visual patterns (\textit{e.g.}, spatial structures and textures) when the source and target domains are closely related. 
4)~\texttt{deb}LoRA consistently outperforms long-tailed recognition methods, ResLT~\citep{cui2022reslt} and SADE~\citep{zhang2022self} (2.5 and 2.9 points by average).
ResLT and SADE mainly introduce re-weighting strategies to balance the learning of different classes, but they do not directly rectify the bias in the feature space.
In contrast, \texttt{deb}LoRA explicitly learns a de-biased representation center for tail classes.
5)~We further validate the generalizability of our method by conducting experiments on additional long-tailed datasets Places365-LT~\citep{liu2019large}, iNaturalist~\citep{van2018inaturalist}, and fMoW-S2~\citep{christie2018functional,cong2022satmae}. Our \texttt{deb}LoRA consistently outperforms baselines, achieving up to 7.2\% improvement on tail classes (see Appendix).
2)~\textit{Oriented Object Detection.}
We validate our method's generalization ability on the oriented object detection task in Table~\ref{tab:comp_det}.
We have two key findings.
1) Our \texttt{deb}LoRA achieves the highest mAP scores across all positions. Notably, \texttt{deb}LoRA outperforms vanilla LoRA by an impressive 6.7 percentage points.
2) Notably, all methods performed better in the middle classes than in the head. This might be attributed to the greater intra-class variation in head classes, whereas middle classes have more distinct and compact features.

\vspace{-1em}
\section{Conclusion}\label{sec:conclusion}

\vspace{-.3cm}
In this paper, we propose \texttt{deb}LoRA, a novel approach for adapting foundation models to data-scarce and long-tailed remote sensing domains while mitigating representation bias.
Our method introduces unsupervised clustering to capture robust visual attributes shared across classes, and feature calibration to rectify the bias in tail class representations.
We validate the effectiveness of \texttt{deb}LoRA through extensive experiments on multiple RS adaptation settings and downstream tasks, where it consistently outperforms vanilla LoRA and other long-tailed recognition methods.
Notably, \texttt{deb}LoRA achieves significant performance gains on tail classes without sacrificing the performance on head classes, highlighting its ability to learn debiased feature representations.

\vspace{-1em}
\begin{ack}
\vspace{-1em}
The author gratefully acknowledges the support from the DSO research grant awarded by DSO National Laboratories, Singapore, and the Lee Kong Chian Fellow grant awarded to Dr.\ Qianru Sun by Singapore Management University.
\end{ack}

\newpage
\clearpage
\bibliographystyle{plainnat}
\bibliography{reference}

\newpage
\clearpage

\appendix

\newcounter{apptable}
\newcounter{appfigure}

\renewcommand{\thetable}{A\arabic{apptable}}
\renewcommand{\thefigure}{A\arabic{appfigure}}

\let\oldwraptable\wraptable%
\let\endoldwraptable\endwraptable%
\renewenvironment{wraptable}[2]{%
    \refstepcounter{apptable}%
    \oldwraptable{#1}{#2}%
    }{%
    \endoldwraptable%
}

\let\oldcaption\caption%
\renewcommand{\caption}{%
    \ifx\@captype\table\else\refstepcounter{appfigure}\fi%
    \oldcaption%
}

\section{Appendix}\label{sec_appendix}
This appendix contains the following supplementary information:
\begin{enumerate}
\item Section~\ref{sec:fusrs} details on the customized SAR ship dataset used in the ORS $\to$ SAR setting, complementing the experiments in Section~\ref{sec:exp}.
\item Section~\ref{sec:additional_datasets} presents experiments on additional datasets, including natural image datasets and a multi-spectral remote sensing dataset, to demonstrate the generalizability of our method.
\item Section~\ref{sec:ablation_analysis} provides ablation studies and additional analyses, including quantitative feature analysis, sensitivity to cluster number $K$, and statistical analysis with error bars.
\item Section~\ref{sec:limitations} discusses the limitations of our work.
\end{enumerate}

\subsection{Details of the customized SAR ship dataset}\label{sec:fusrs}
\begin{figure}[htbp]
    \centering
    \includegraphics[width=0.95\linewidth]{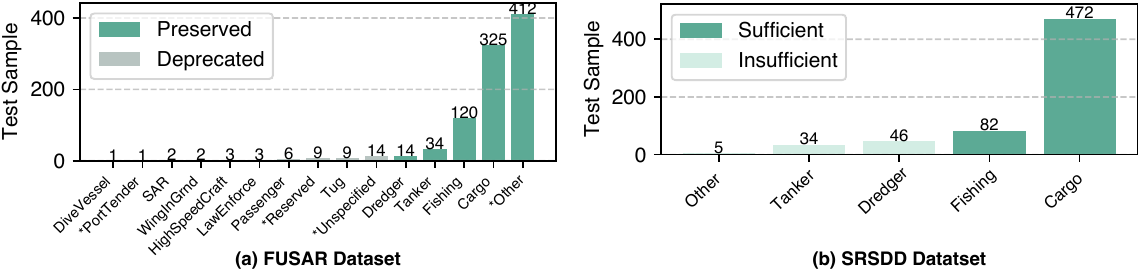}
    \vspace{-.8em}
    \caption{\textbf{Constraints of the SAR datasets' test sets.} This figure illustrates the per-category test sample distribution of (a) the FUSAR dataset and (b) the SRSD dataset. The FUSAR dataset suffers from insufficient test samples and vaguely defined classes (indicated by ``$\ast$''). Similarly, the SRSDD dataset also has the issue of insufficient test samples.}\label{fig:fusar}
\end{figure}

We selected the FUSAR-Ship~\cite{hou2020fusar} and SRSDD~\cite{lei2021srsdd} datasets as our source datasets due to their high resolution ($\leq 10$m) and fine-grained ship subcategories, as shown in Figure~\ref{fig:fusar}. However, both datasets have limitations. Figure~\ref{fig:fusar}(a) shows that the FUSAR-Ship dataset has insufficient test samples (i.e., certain categories have only $\leq15$ test samples) and unclear category definitions (e.g., ``Reserved'' or ``Unspecified'' categories). Figure~\ref{fig:fusar}(b) reveals that the SRSDD dataset also suffers from insufficient test samples. To address these issues and establish a robust benchmark, we combined the ship categories from both datasets, merging those with fewer than 10 test samples into an ``others'' category.

\subsection{Experiments on Additional Datasets}\label{sec:additional_datasets}
To demonstrate the generalizability of our \texttt{deb}LoRA method, we conducted experiments on three additional datasets: two from the natural image domain (Places365-LT~\cite{liu2019large} and iNaturalist 2018~\cite{van2018inaturalist}) and one multi-spectral remote sensing dataset (fMoW-S2~\cite{christie2018functional,cong2022satmae}). These datasets were chosen for their unique properties: 
1) Places365-LT exhibits a substantial domain gap from Stable Diffusion's pre-training data, allowing us to evaluate the performance of our domain adaptation model.
2) iNaturalist 2018 has a high imbalance ratio of 500, enabling us to assess our model's performance under severe class imbalance conditions.
3) fMoW-S2 contains multi-spectral data, including visible, near-infrared, and shortwave infrared bands, complementing our existing experiments on optical (DOTA) and SAR (FUSRS) imagery.
The results are given in Table~\ref{tab:natural_datasets} and Table~\ref{tab:fmow_results}.

\begin{wraptable}{r}{0.5\textwidth}
    \vspace{-1.9em}
    \caption{\textbf{Comparison on Places365-LT and iNaturalist2018 datasets}.
    Results reported in top-1 accuracy (\%).
    Our results are marked in \setlength{\fboxsep}{2.5pt}\colorbox{gray!20}{gray}. 
    }\label{tab:natural_datasets}
    \vspace{-.2em}
    \resizebox{\linewidth}{!}{%
    \begin{tabular}{lcccccccccccc}
    \toprule
    \multirow{2}{*}{\textbf{Method}} & \multicolumn{3}{c}{\textbf{Places365-LT}} & \multicolumn{3}{c}{\textbf{iNaturalist 2018}} & \multicolumn{3}{c}{\textbf{Mean}} \\
    \cmidrule(lr){2-4} \cmidrule(lr){5-7} \cmidrule(lr){8-10}
    & Head & Middle & Tail & Head & Middle & Tail & Head & Middle & Tail \\
    \midrule \midrule
    Zero-Shot & 40.3 & 36.9 & 24.9 & 36.2 & 29.4 & 8.9 & 38.3 & 33.2 & 16.9 \\
    Fine-Tune & 43.2 & 31.1 & 39.0 & 66.5 & 69.2 & 67.5 & 54.9 & 50.2 & 53.3 \\
    LoRA & 48.2 & 42.0 & 44.9 & 71.9 & 74.6 & 71.2 & 60.1 & 58.3 & 58.1 \\
    \rowcolor{gray!20}  \quad w/ \texttt{deb}LoRA & \textbf{50.9} & \textbf{51.2} & \textbf{49.2} & \textbf{72.6} & \textbf{79.8} & \textbf{78.4} & \textbf{61.8} & \textbf{65.5} & \textbf{63.8} \\
    \bottomrule
    \end{tabular}
    }
    \vspace{-1.8em}
\end{wraptable}
1) On Places365-LT and iNaturalist 2018 (Table~\ref{tab:natural_datasets}), \texttt{deb}LoRA consistently outperforms LoRA, especially for tail classes. We observe improvements of 4.3\% and 7.2\% for Places365-LT and iNaturalist 2018 tail classes, respectively.

2) For the fMoW-S2 dataset (Table~\ref{tab:fmow_results}), we adapted Stable Diffusion (SD) to the scene recognition task. The dataset was manually divided into ``Head'' (34 classes comprising 80\% of the

\begin{wraptable}{r}{0.3\textwidth}
    \vspace{-.7em}
    \caption{\textbf{Results on the fMoW-S2 dataset.}}\label{tab:fmow_results}
    \resizebox{\linewidth}{!}{%
    \begin{tabular}{lccc}
    \toprule[1.3pt]
    \multirow{2}{*}{\textbf{Method}} & \multicolumn{3}{c}{\textbf{SD $\to$ fMoW-S2}} \\
    \cmidrule(lr){2-4}
    & Head & Tail & Overall \\
    \midrule \midrule
    Fine-Tune & 46.2 & 34.6 & 44.9 \\
    \midrule
    LoRA & 46.5 & 38.1 & 46.2 \\
    \quad w/ ResLT & 46.8 & 38.6 & 46.5 \\
    \rowcolor{gray!20} \quad w/ \texttt{deb}LoRA & \textbf{46.8} & \textbf{41.2} & \textbf{46.8} \\
    \bottomrule[1.3pt]
    \end{tabular}
    }
    \vspace{-1em}
\end{wraptable}
samples) and ``Tail'' (28 classes comprising 20\% of the samples). Results were evaluated by linear probing.
\texttt{deb}LoRA achieves the highest overall accuracy (46.8\%) and tail class accuracy (41.2\%), surpassing the second-best method (ResLT) by 0.3 and 2.6 percentage points, respectively.

These results confirm that our method effectively addresses the long-tailed distribution problem across various domains, including natural images and multi-spectral remote sensing data. The consistent improvements, particularly for tail classes, highlight the robustness of \texttt{deb}LoRA in handling class imbalance and domain adaptation challenges.

\subsection{Ablation Studies and Additional Analyses}~\label{sec:ablation_analysis}
To provide a comprehensive evaluation of our \texttt{deb}LoRA method, we conducted several ablation studies and additional analyses. These experiments aim to validate the effectiveness of our approach, investigate its sensitivity to key hyperparameters (\textit{i.e.}, cluster number $K$), and demonstrate the statistical significance.

\begin{wraptable}{r}{0.45\textwidth}
    \vspace{-1.9em}
    \caption{
    \textbf{Quantitative feature analysis on the DOTA dataset.}
    Inter-class distance is measured as the average cosine distance between class centers, while intra-class distance is the average cosine distance between samples and their corresponding class centers.
    }\label{tab:feature_analysis}
    \resizebox{\linewidth}{!}{%
    \begin{tabular}{lccc}
    \toprule
    \multirow{2}{*}{\textbf{Method}} & \multicolumn{2}{c}{\textbf{Inter-class}} & \textbf{Intra-class} \\
    \cmidrule(lr){2-3}\cmidrule(lr){4-4}
    & Head-Tail & Tail-Tail & Tail \\
    \midrule
    Fine-tuning    & 0.674 & 0.621 & 0.170 \\
    LoRA           & 0.702 & 0.607 & 0.182 \\
    \rowcolor{gray!16} \quad w/ \texttt{deb}LoRA & 0.719 & 0.632 & 0.146 \\
    \bottomrule
    \end{tabular}
    }
    \vspace{-1.3em}
\end{wraptable}
\noindent\textbf{Quantitative Feature Analysis.} To further validate the effectiveness of our \texttt{deb}LoRA method, we present a quantitative analysis of the learned features, focusing on inter-class and intra-class distances. Table~\ref{tab:feature_analysis} shows the results on the DOTA dataset.

Our analysis reveals several key observations about \texttt{deb}LoRA. First, it enlarged the inter-class distance between tail and head classes, with the average cosine distance increasing from 0.702 to 0.719. This indicates improved separation between these class groups. Second, \texttt{deb}LoRA reduced the intra-class distance for tail classes, as evidenced by the decrease in average cosine distance from 0.182 to 0.146. This suggests a tighter clustering of tail samples. Finally, we observed an increase in inter-class distance among tail classes, with the average cosine distance rising from 0.607 to 0.632. This demonstrates better separation among different tail classes. These findings support the effectiveness of \texttt{deb}LoRA in improving feature separation for tail classes.

\begin{wraptable}{r}{0.29\textwidth}
    \vspace{-1.9em}
    \caption{\textbf{Ablation study on the number of clusters ($K$) in debLoRA.} Our default value is marked in \setlength{\fboxsep}{2.5pt}\colorbox{gray!20}{gray}.}\label{tab:k_ablation}
    \resizebox{\linewidth}{!}{%
    \begin{tabular}{lcccc}
    \toprule[1.3pt]
    \multirow{2}{*}{\textbf{$K$}} & \multicolumn{3}{c}{\textbf{Macro F1 Score (\%)}} \\
    \cmidrule(lr){2-4}
    & Head & Middle & Tail \\
    \midrule \midrule
    16 & 99.1 & 96.9 & 90.4 \\
    \rowcolor{gray!20}  32 & 99.3 & 97.7 & 95.1 \\
    64 & 99.3 & 97.4 & 94.8 \\
    \bottomrule[1.3pt]
    \end{tabular}
    }
    \vspace{-2em}
\end{wraptable}
\noindent\textbf{Sensitivity to Cluster Number $K$.} We conducted an ablation study to investigate the sensitivity of our method to the number of clusters (K) used in the de-biasing process. Table~\ref{tab:k_ablation} shows the results on the SD $\to$ DOTA adaptation.

From the table we can observe that performance generally improves as $K$ increases, with the most significant gains observed for tail classes. For instance, when $K$ increases from 16 to 32, the F1 score for tail classes improves by 4.7\%. The performance peak around $K$=32 suggests a good default value for our method. These findings indicate that our method is sensitive to $K$ but remains effective across different values.

\begin{wraptable}{r}{0.5\textwidth}
    \vspace{-1.9em}
    \caption{\textbf{Error Bar Analysis.} Reported in mean $\pm$ std. Our results are marked in \setlength{\fboxsep}{2.5pt}\colorbox{gray!20}{gray}.}\label{tab:error_bars}
    \resizebox{\linewidth}{!}{%
    \begin{tabular}{lccc}
    \toprule[1.3pt]
    \multirow{2}{*}{\textbf{Method}} & \multicolumn{3}{c}{\textbf{Macro F1 Score (\%)}} \\
    \cmidrule(lr){2-4}
    & Head & Middle & Tail \\
    \midrule \midrule
    Zero-Shot & 99.2 $\pm$ 0.1 & 97.4 $\pm$ 0.3 & 87.6 $\pm$ 0.6 \\
    Fine-Tune & 99.1 $\pm$ 0.1 & 96.7 $\pm$ 0.1 & 86.8 $\pm$ 0.2 \\
    \midrule
    LoRA & 99.3 $\pm$ 0.1 & 97.2 $\pm$ 0.1 & 91.8 $\pm$ 0.2 \\
    \quad w/ ResLT & 99.3 $\pm$ 0.1 & 97.5 $\pm$ 0.3 & 92.9 $\pm$ 0.3 \\
    \rowcolor{gray!20} \quad w/ \texttt{deb}LoRA & \textbf{99.3 $\pm$ 0.1} & \textbf{97.5 $\pm$ 0.2} & \textbf{94.8 $\pm$ 0.3} \\
    \bottomrule[1.3pt]
    \end{tabular}
    }
    \vspace{-1.3em}
\end{wraptable}
\noindent\textbf{Statistical Analysis with Error Bars.} To demonstrate the statistical significance of our results, we report the results of three runs with random initializations on the SD $\to$ DOTA experiment. Table~\ref{tab:error_bars} shows the results.

These results demonstrate that our \texttt{deb}LoRA method consistently outperforms other approaches, especially for tail classes, with statistically stable improvements. The small std across all methods indicate the stability of the results. Notably, \texttt{deb}LoRA shows the most substantial improvement for tail classes, with a mean F1 score of 94.8\% and a standard deviation of only 0.3\%, highlighting both the effectiveness and consistency of our approach in addressing the long-tailed distribution problem.

\subsection{Limitations}\label{sec:limitations}
While our proposed \texttt{deb}LoRA method has proven effective in adapting foundation models to remote sensing domains with limited data and long-tailed distributions, we acknowledge three key limitations:
\noindent \textbf{Assumption of shared visual attributes.} Our method assumes that visual attributes are shared across classes, enabling robust representation learning through clustering. However, if the visual attributes are highly class-specific or there is significant intra-class variation, the effectiveness of our approach may be reduced.

\noindent \textbf{Sensitivity to hyperparameters.} The performance of \texttt{deb}LoRA depends on the selection of hyperparameters, such as the number of clusters $K$. The optimal value of $K$ may differ depending on the specific dataset and adaptation setting.

\noindent \textbf{Limited evaluation on SAR datasets.} Due to the scarcity of large-scale SAR datasets with sufficient samples for reliable evaluation, we created a customized dataset by combining two existing SAR datasets. Further investigation is needed to assess the performance of our method on a broader range of SAR datasets and tasks.

By acknowledging these limitations, we aim to provide a transparent and objective assessment of our work and to encourage future research addressing these challenges to further improve long-tailed adaptation in remote sensing domains.

\end{document}